\documentclass[journal,twoside,web]{ieeecolor}
\usepackage{lcsys}
\usepackage{cite}
\usepackage{amsmath,amssymb,amsfonts}
\usepackage{graphicx}
\usepackage{textcomp}
\def\BibTeX{{\rm B\kern-.05em{\sc i\kern-.025em b}\kern-.08em
    T\kern-.1667em\lower.7ex\hbox{E}\kern-.125emX}}
\usepackage{amsfonts}
\usepackage{mathrsfs}  
\usepackage{times}
\usepackage{bm}
\usepackage[dvipsnames]{xcolor}
\usepackage{amsmath,amssymb,latexsym}
\usepackage{graphicx}
\usepackage{multirow}
\usepackage{hyperref}
\usepackage{booktabs, makecell, tabularx, tabulary}

\usepackage{dsfont}
\usepackage{epstopdf}
\usepackage{caption}
\usepackage{subcaption}
\usepackage{algorithm}
\usepackage{algpseudocode}
\usepackage{cite}

\newtheorem{definition}{Definition}

\newtheorem{remark}{Remark}

\newtheorem{theorem}{Theorem}

\title{Robust Classification using Contractive Hamiltonian Neural ODEs}

\author{Muhammad Zakwan, Liang Xu, \IEEEmembership{Member, IEEE}, and Giancarlo Ferrari-Trecate, \IEEEmembership{Senior Member, IEEE}
\thanks{This research is supported by the Swiss National Science Foundation under the NCCR Automation (grant agreement 51NF40 180545). (Corresponding author: Liang Xu.)}
\thanks{Authors are with the Institute of Mechanical Engineering, Ecole Polytechnique Fédérale de Lausanne (EPFL), CH-1015 Lausanne, Switzerland {\tt\small \{muhammad.zakwan, liang.xu, giancarlo.ferraritrecate\}@epfl.ch }}%
}

\begin{document}

\maketitle
\thispagestyle{empty}
\pagestyle{empty}

\begin{abstract}
Deep neural networks can be fragile and sensitive to small input perturbations that might cause a significant change in the output. 
In this paper, we employ contraction theory to improve the robustness of neural ODEs (NODEs). 
A dynamical system is contractive if all solutions with different initial conditions converge to each other exponentially fast.
As a consequence, perturbations in initial conditions become less and less relevant over time.
Since in NODEs the input data corresponds to the initial condition of dynamical systems, we show contractivity can mitigate the effect of input perturbations. 
More precisely, inspired by NODEs with Hamiltonian dynamics, we propose a class of contractive Hamiltonian NODEs (CH-NODEs). 
By properly tuning a scalar parameter, CH-NODEs ensure contractivity by design and can be trained using standard backpropagation.
Moreover, CH-NODEs enjoy built-in guarantees of non-exploding gradients, which ensure a well-posed training process. 
Finally, we demonstrate the robustness of CH-NODEs on the MNIST image classification problem with noisy test data.
\end{abstract}

\begin{IEEEkeywords}
Machine learning,  Neural networks, Stability of nonlinear systems
\end{IEEEkeywords}
\section{INTRODUCTION}


\IEEEPARstart{N}{eural} networks (NNs) have demonstrated stunning performance in image classification, natural language processing, and speech recognition tasks.
However, they can be sensitive to input noise or adversarial attacks~\cite{RN11030}.
For example, in image classification problems using convolutional NNs, it has been shown that small perturbations to input images can lead to misclassification~\cite{RN11033}.
The common solutions are either heuristic, for example, feature obfuscation~\cite{RN11002} and adversarial training~\cite{RN11056}, or certificate-based such as Lipschitz regularization \cite{RN11030,RN11084,RN11751}.
The general purpose of certificate-based techniques is to penalize the input-to-output sensitivity of a NN to improve robustness.
To do so, they usually require layerwise regularization, which is computationally expensive.

Recently, the connections between NNs and dynamical systems have been extensively explored.
Representative results include classes of NNs stemming from the discretization of dynamical systems~\cite{RN10726} and Neural ODEs (NODEs)~\cite{RN10739}.
Instead of specifying a discrete sequence of hidden layers, NODEs transform the input through a continuous-time ODE embedding training parameters. 
The continuous-time nature of NODEs makes them particularly suitable for learning complex dynamical systems~\cite{RN11780, RN10758} and allows borrowing tools from the system theory to analyze NN properties.

Although NODEs have shown greater robustness against random perturbations than common convolutional NNs~\cite{RN11745}, research on methods guaranteeing more robustness is still very active.
For instance,  to improve resilience to adversarial attacks, NODEs equipped with Lyapunov-stable equilibrium points have been proposed~\cite{RN11716}. 
Likewise, \cite{rodriguez2022lyanet} introduces a loss function to promote robustness based on a control-theoretic Lyapunov condition.
Both methods have shown promising performance against
adversarial attacks. Nonetheless, they do not provide robustness
guarantees \emph{by design}. The authors in~\cite{massaroliStableNeuralFlows2020} design provably stable NODE and argue that stability can reduce the sensitivity to small
perturbations of the input data. Nevertheless, the claim is not supported by theoretical analysis or numerical validation.

In this paper, we employ contraction theory to enhance the robustness of NODEs. A dynamical system is contractive if all trajectories converge to each other exponentially fast~\cite{RN11742, tsukamoto2021contraction}.
From the lens of contraction, slight perturbations of initial conditions have a diminishing impact over time on the NODE state.
We highlight that the notion of contractivity allows one to deal with non-autonomous systems while studying the stability of multiple trajectories at once.
Contractivity has been used to improve the well-posedness of implicit neural networks \cite{RN11770} and the trainability of recurrent neural networks~\cite{RN11159}. Nevertheless, to the authors' knowledge, the relation between contractivity and robustness of NODEs has not been explored.

With the above considerations, we propose a class of NODEs with built-in contraction properties based on the theory of Hamiltonian systems, which we refer to as contractive Hamiltonian NODEs (CH-NODEs). Hamiltonian systems are structured nonlinear systems that can model complex dynamics~\cite{RN10727}.
Since their built-in dissipation properties facilitate contraction analysis, they are natural candidates for building contractive NODEs.  
Moreover, CH-NODEs do not rely on stability-promoting regularizers/costs~\cite{rodriguez2022lyanet,RN11716}, which can increase the training complexity, hence limiting the depth of the NNs that can be used in applications.

In general, NODEs rely on gradient based algorithms to update weights. The appearance of vanishing or exploding gradient during training may cause numerical instability, thus compromising the learning process.
We however prove that CH-NODEs cannot have exploding gradients, irrespective of their depth and for all possible values of the parameters.
Finally, we test CH-NODEs in benchmark classification problems and show that they can achieve substantially better accuracy than alternative architectures on test dataset corrupted by noise.

{\it Notation:} The set of non-negative real numbers is $\mathbb{R}^+$. The operator $\frac{\partial^2 f(\bm{x})}{\partial \bm{x}^2}$ denotes the Hessian of a continuously differentiable function $f(\cdot)$. The maximal and minimal eigenvalues of a matrix $\bm{A}$ are denoted as $\bar{\lambda}(\bm{A})$ and $\underline{\lambda}(\bm{A})$, respectively. $\text{diag}(\bm{x})$ represents a diagonal matrix with the entries of the vector $\bm{x}$ on the diagonal. For symmetric matrices $\bm{A}$, and $\bm{B}$, $\bm{A} \succ (\succeq) \bm{B}$ means that $\bm{A}-\bm{B}$ is positive (semi)definite. The 2-norm is denoted as $||\cdot||$.

{\it Organization:} 
This paper is organized as follows:
Section II provides the preliminaries.
In Section III, we introduce and analyze CH-NODEs.
The performance evaluation is conducted in Section IV and Section V concludes the paper.

\section{PRELIMINARIES}

We briefly introduce NODEs and describe how contractivity can improve their robustness. Moreover, we recall a few preliminaries related to Hamiltonian systems.
\subsection{Neural ODEs}
A NODE~\cite{RN10739} is a continuous-depth model defined by 
\begin{align}
\label{eq:1}
   & \bm{\xi}(0) = \phi(\bm{x}, \omega )\;,\;  (\text{input layer}) \\ 
&\label{eq:2}    \dot{\bm{\xi}}(t) = f_t(\bm{\xi}(t), \theta(t))\;,\;  (\text{continuum of hidden layers}) \\ 
&\label{eq:3}     \bm{y}(T) = \psi(\bm{\xi}(T), \eta)\;,\;  (\text{output layer})
\end{align}
where $\bm{x} \in \mathbb{R}^{m}$ is the input data (e.g an image), $\bm{\xi} \in \mathbb{R}^n$ is the state, and $\phi, f, \psi$ are neural networks with trainable parameters $\omega, \theta, \eta$, respectively. The dynamics \eqref{eq:2} evolves over the time interval $[0, T]$ and $\bm{y}(T)$ is the output.
The training of NODEs \eqref{eq:1}-\eqref{eq:3} refers to minimizing a given loss function over trainable parameters.
\begin{remark}
\label{remark_discretization}
To specify a NN architecture for the training, one can discretize the ODE \eqref{eq:2}
with a sampling period $h = \frac{T}{N}$, $N \in  \mathbb{N}$. The $N$
resulting discrete-time equations can be utilized to define the layers of a NN of depth $N$ \cite{RN10739}. For instance, using Forward Euler (FE) discretization with a step-size of $h > 0$, one obtains 
\begin{align}
\label{eq:FE}
    \bm{\xi}_{i + 1 } = \bm{\xi}_i + h f_i(\bm{\xi}_i, \theta_i), \ \text{for} \ i = 0,1,\cdots, N-1 \; ,
\end{align}
 where $\bm{\xi}_i$ and $\bm{\xi}_{i+1}$ represent the input and output of layer $i$, respectively. In practice, the step-size $h$ is chosen sufficiently small so as to interpret the states in~\eqref{eq:FE} as a sampled version of the state $\bm{\xi}(t)$ in~\eqref{eq:2}.
\end{remark}


Neural ODEs are not necessarily robust, as input perturbations can produce a significant change in outputs~\cite{RN11745}.
In the following, we introduce  contraction theory for dynamical systems and show how it can be used for improving the robustness of NODEs.
\subsection{Contraction theory}
\begin{definition}[\cite{RN11742}]
\label{Def:1}
Let $\bm{\xi}(t)$ and $\tilde{\bm{\xi}}(t)$ be two solutions of \eqref{eq:2} starting from  $\bm{\xi}(0)$ and $\tilde{\bm{\xi}}(0)$, respectively. Then~\eqref{eq:2} is contractive if $ \exists C >0$, and a contraction rate $\rho > 0$, such that
 \begin{align}
 \label{Contract_def}
|| \tilde{\bm{\xi}}(t) - \bm{\xi}(t)|| \leq  C e^{- \rho t} || \tilde{\bm{\xi}}(0) - \bm{\xi}(0) ||     
 \end{align}
for all $t > 0$ and $\bm{\xi}(0), \tilde{\bm{\xi}}(0) \in \mathbb{R}^n$.
\end{definition}

According to \eqref{Contract_def}, perturbations in the input data of contractive NODEs vanish exponentially fast as $t$ increases.
\begin{remark}
\label{remark1}
   As shown in \cite{RN11742}, all the state trajectories of a contractive NODE converge exponentially to an equilibrium trajectory. This may limit the flexibility of NODEs. However, a loss of expressivity seems unavoidable for increasing robustness, as discussed in  \cite{tsipras2018robustness}. 

\end{remark}



\subsection{Hamiltonian System Theory}

A Hamiltonian system~\cite{RN10727} is described by
\begin{align}
\label{eq.Plant}
  \begin{aligned}
    \dot{\bm{\xi}}(t) &= (\bm{J}(\bm{\xi}(t))-\bm{R}(\bm{\xi}(t)))  \displaystyle{ \frac{\partial {H}(\bm{\xi}(t),t)}{\partial \bm{\xi}(t)}} \;,  \end{aligned}
\end{align}
where 
$\bm{J}(\bm{\xi}(t)) = -\bm{J}^\top(\bm{\xi}(t))$, and $\bm{R}(\bm{\xi}(t)) = \bm{R}^\top(\bm{\xi}(t)) \succeq 0$ are interconnection and damping matrices, respectively.
The continuously differentiable function $H: \mathbb{R}^{n} \times \mathbb{R}^+\rightarrow \mathbb{R}$ is called the \emph{Hamiltonian}, and
can be interpreted as the energy of the system~\eqref{eq.Plant}.
It is straightforward to show that, $$\dot{H}(t)=- \frac{\partial {H}^\top(\bm{\xi}(t),t)}{\partial \bm{\xi}(t)}\bm{R}(\bm{\xi}(t)) \frac{\partial {H}(\bm{\xi}(t),t)}{\partial \bm{\xi}(t)}\le 0 \;.$$
Therefore, the system~\eqref{eq.Plant} dissipates energy and $\bm{R}(\xi(t))$ determines the dissipation rate.
In the next section, we will illustrate how to exploit these properties to guarantee contractivity.

\section{Contractive Hamiltonian Neural ODEs}

We introduce CH-NODEs and show that their parameterization allows one to perform training  using standard gradient based algorithms.  Moreover, we show that CH-NODEs enjoy built-in guarantees of non-exploding gradients.
Consider the following NODE~\footnote{We omit input and output layers for brevity. They are usually selected depending on the specific learning task \cite{RN10739}.}
\begin{align}
\label{general_ODE}   
\begin{aligned}
            \dot{\bm{\xi}}(t) &= \bm{F}(t,\bm{\xi}(t))(\bm{K}^\top(t) \sigma{(\bm{K}(t)\bm{\xi}(t) + \bm{b}(t))} \\ &\hspace{10mm}+ (\bm{L}^\top(t) \bm{L}(t) +\kappa \bm{I} )\bm{\xi}(t) ), \  t \in [0, \ T], 
\end{aligned}
\end{align}
where $\bm{F}$ is a given matrix-valued function, $\kappa>0$ is a constant, $\bm{K}: \mathbb{R}^+ \rightarrow \mathbb{R}^{n \times n}$, $\bm{b}: \mathbb{R}^+ \rightarrow \mathbb{R}^{n}$, and $\bm{L}: \mathbb{R}^+ \rightarrow \mathbb{R}^{n \times n}$ are trainable parameters.
The map $\sigma : \mathbb{R} \rightarrow \mathbb{R}$ is the \emph{activation function} and is  applied element-wise when the argument is a matrix. Moreover, we assume $\sigma(\cdot)$ is differentiable almost everywhere (e.g. $\tanh{(\cdot)}$, softmax$(\cdot)$) and satisfies
$0 \leq  {\sigma'}(x) \leq S$,
for some $S > 0$, where ${\sigma'}(x)$ denotes any sub-derivative.
One can easily verify that~\eqref{general_ODE} is a Hamiltonian system when $\bm{F}(t,\bm{\xi}(t)) = \bm{J}(\bm{\xi}(t))- \bm{R}(\bm{\xi}(t))$ for some anti-symmetric matrix $\bm{J}(\bm{\xi}(t))$ and  $\bm{R}(\bm{\xi}(t)) \succeq 0$.
Indeed, in such case, by choosing the Hamiltonian in~\eqref{eq.Plant} as 
\begin{align}
\label{energy_func}
\begin{aligned}
    H(\bm{\xi}(t), t) &=  \tilde{\sigma}(\bm{K}(t)\bm{\xi}(t) + \bm{b}(t))^\top \mathds{1}_n  \\ 
    & \hspace{-2mm}+ \frac{1}{2} \bm{\xi}^\top(t) (\bm{L}^\top(t) \bm{L}(t) +\kappa \bm{I}) \bm{\xi}(t), \ t \in [0,  T],
\end{aligned}
\end{align}
where $\tilde{\sigma} : \mathbb{R} \rightarrow \mathbb{R}$ is a differentiable map satisfying $\tilde{\sigma}'(\cdot) = \sigma(\cdot)$, and $\mathds{1}_n$ represents a column vector with $n$ elements equal to $1$, one obtains the ODE \eqref{general_ODE}.
\begin{remark}
The dynamics in~\eqref{general_ODE} encompasses several standard NODEs. For instance, if $\bm{F}(t,\bm{\xi}(t))$ is an anti-symmetric matrix, $\bm{L}(t) = 0$, and $\kappa = 0$, we recover the Hamiltonian Deep Neural Networks (H-DNNs) discussed in~\cite{RN11564}. It is shown in~\cite{RN11564} that H-DNNs unifies several important architectures proposed in~\cite{RN10726} and \cite{changAntisymmetricrnnDynamicalSystem2019}. Moreover, by setting  $\bm{F}(t,\bm{\xi}(t)) = \bm{K}^{-1}(t)$, $\bm{L}(t) = 0$, and $\kappa = 0$, and using FE discretization we recover the layer equation \eqref{eq:FE} of a class of NNs called residual networks (ResNets). We highlight that several classes of ResNets enjoy universal approximation properties \cite{linResNetOneneuronHidden2018}. 
\end{remark}
\subsection{Contractive Hamiltonian Neural ODEs}
We show that for a suitable selection of parameters in~\eqref{general_ODE}, one obtains a contractive NODE by design.
The following theorem, proven in Appendix \ref{appen1}, provides this result.

\begin{theorem}
  \label{prop-1}
  For a given constant skew-symmetric matrix $\bm{J}$, let $\bm{F}$ in~\eqref{general_ODE} be equal to $\bm{J}-\gamma \bm{I}$ and define
\begin{align}
 \hspace{-3mm}c_1 &=  \inf_{s \in [0, T]} \underline{\lambda}(\bm{L}^\top(s)\bm{L}(s)) + \kappa\; , \label{eq.c1Definition}\\
 \hspace{-3mm}  c_2 &= \sup_{s \in [0, T]} (\bar{\lambda}(\bm{L}^\top(s)\bm{L}(s)) + S\bar{\lambda}(\bm{K}^\top(s) \bm{K}(s))) + \kappa\;, \hspace{-2mm} \label{eq.c2Definition} 
\end{align}
and $  \alpha = \displaystyle{\frac{c_2 - c_1}{c_2 + c_1}}$. If $\epsilon>0$ is such that $1-\alpha^2-\epsilon>0$ and $\gamma$ verifies
 \begin{align}
\displaystyle{\gamma \geq \sqrt{\frac{(\alpha^2 + \epsilon)\bar{\lambda}(\bm{J} \bm{J}^\top)}{1 - \alpha^2 - \epsilon}}}\;,\label{eq:gammaCondition}
\end{align}
then, for $t \in [0, \ T]$, the NODE~\eqref{general_ODE} becomes
\begin{align}
\label{CH_NODE}   
\begin{aligned}
            \dot{\bm{\xi}}(t) &= (\bm{J} - \gamma \bm{I})(\bm{K}^\top(t) \sigma{(\bm{K}(t)\bm{\xi}(t) + \bm{b}(t))} \\ &\hspace{20mm}+ (\bm{L}^\top(t) \bm{L}(t) +\kappa \bm{I} )\bm{\xi}(t) ) \ ,   
\end{aligned}
\end{align}
and is contractive with a contraction rate $\epsilon$.
\end{theorem}

The essence of Theorem \ref{prop-1} lies in scaling the dissipation term $\gamma$ as in \eqref{eq:gammaCondition},
such that \eqref{CH_NODE} remains contractive for arbitrary weights.
However, the selection of $\gamma$ relies on NODE parameters $\bm{K}(t), \bm{L}(t)$, which are changing during the training.
Nevertheless, the following iterative procedure can be adopted to update $\gamma$ such that the trained NODE remains contractive during training: 
i) perform a forward propagation through  \eqref{CH_NODE} with the current value of $\gamma$ and weights and calculate the training loss, ii) update the weights via backpropagation, iii) calculate $c_1$ and $c_2$ for the new weights and update $\gamma$, and $\epsilon$. In the sequel, we refer to the NODE~\eqref{CH_NODE} as CH-NODE.
Although we do not have formal guarantees on the convergence of the iterative algorithm, it never failed to converge in numerical experiments, including those presented in Section IV. Moreover, early stopping of the algorithm at any iteration provides a contractive CH-NODE.

\subsection{Non-exploding gradients}
From the implementation point of view, neural ODEs may have  vanishing or exploding gradients during training, which may cause numerical instability. These phenomena are related to the convergence to zero or the divergence, respectively, of the Backward Sensitivity Matrices (BSMs) arising in gradient computations through backpropagation \cite{RN11564}.
The continuous-time BSM for the NODE~\eqref{CH_NODE} is defined as
\begin{align}
\label{BSM}
    \frac{\partial \bm{\xi}(T)}{\partial \bm{\xi}(T - t)}, \quad \forall t \in [0, T].
\end{align}
Note that the BSM represents the sensitivity of the output with respect to the intermediate state (hidden continuum) of a NODE.
This term is used to update the NODE parameters during gradient descent \cite{RN11564}.
Therefore, if the norm $||  \frac{\partial \bm{\xi}(T)}{\partial \bm{\xi}(T - t)}||$ is too small or too large, vanishing/exploding gradients may appear and render the learning process unreliable. 

In the following, we show that CH-NODEs~\eqref{CH_NODE} cannot suffer from exploding gradients irrespectively of their depth $T$ and of the parameter values.  The following theorem, proven in Appendix \ref{Appen2}, provides this result.
\begin{theorem}
\label{Thm:2}
The BSM \eqref{BSM} associated with the CH-NODE \eqref{CH_NODE}   satisfies 
\begin{align}
\label{eq:thm2}
   \bigg|\bigg|\frac{\partial \bm{\xi}(T)}{\partial \bm{\xi}(T - t)} \bigg|\bigg| \leq \exp{(- \frac{\rho}{2} t )} \;, 
\end{align}
where $t \in [0, \ T]$, 
$\rho = \frac{\epsilon \beta (\gamma^2 + \bar{\lambda}(\bm{J}\bm{J}^\top))}{ \gamma }$, $\beta = \frac{1}{2}(c_1 + c_2)$, and $\epsilon$ is defined in Theorem \ref{prop-1}.
\end{theorem}
The work \cite{choromanski2020ode} also provides bounds on the norm of the BSMs by enforcing unitary and orthogonality constraints on weight matrices. However, 
 fulfilling them during training may considerably increase the computational burden, whereas we do not impose any constraints on weight matrices. 
 \begin{remark}
By putting $t = T$ in \eqref{eq:thm2}, we get $ ||\frac{\partial \bm{\xi}(T)}{\partial \bm{\xi}(0)} || \leq \exp{(-\frac{\rho}{2} T)} < 1$,
    that is, the input-output sensitivity of the CH-NODE is smaller than $1$. Therefore, a small perturbation at the input would not result in a significant change at the output, i.e., the CH-NODE is robust according to the definition provided in \cite{RN11033}.
\end{remark}
\section{EXPERIMENTS}
In this section, first we show how contractivity improves the robustness of NODEs through a 2D binary classification example. Second, we demonstrate the effectiveness of the CH-NODE \eqref{CH_NODE} on the MNIST image classification task and compare the performance with ResNets \cite{he2016deep}, and  H-DNNs \cite{RN11564}. Third, we numerically validate the non-exploding gradients properties of CH-NODEs\footnote{Further details about the examples and our code are available at \url{https://github.com/DecodEPFL/Contractive-Hamiltonian-Neural-ODEs}}.

\subsection{Contractivity promotes robustness}
\begin{figure}
  \begin{subfigure}[t]{0.5\linewidth}
    \centering
    \includegraphics[width=1.\linewidth]{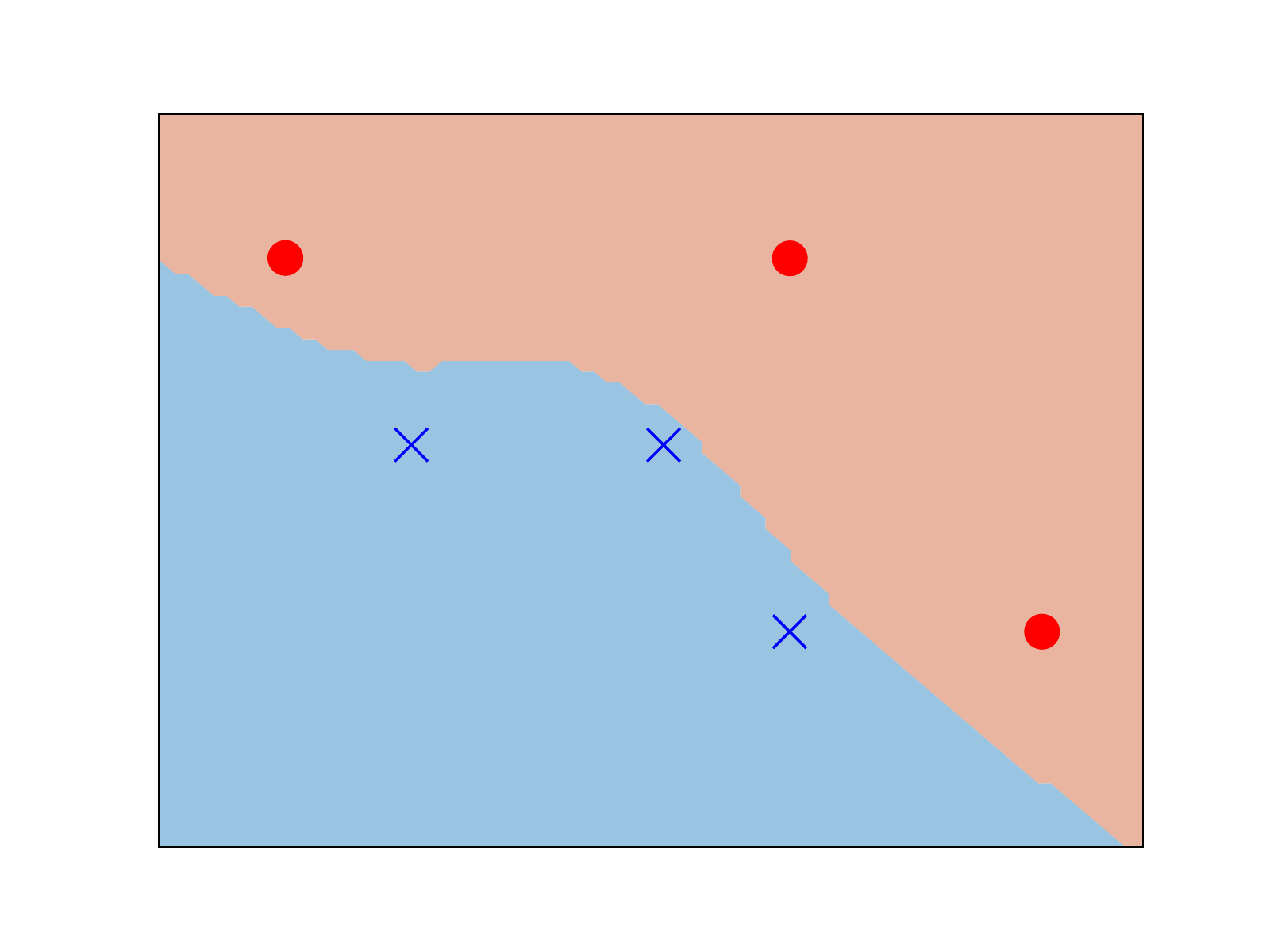} 
    \caption{Classification} 
    \label{fig7:a} 
    \vspace{0.5ex}
  \end{subfigure}
  \begin{subfigure}[t]{0.5\linewidth}
    \centering
    \includegraphics[width=1.\linewidth]{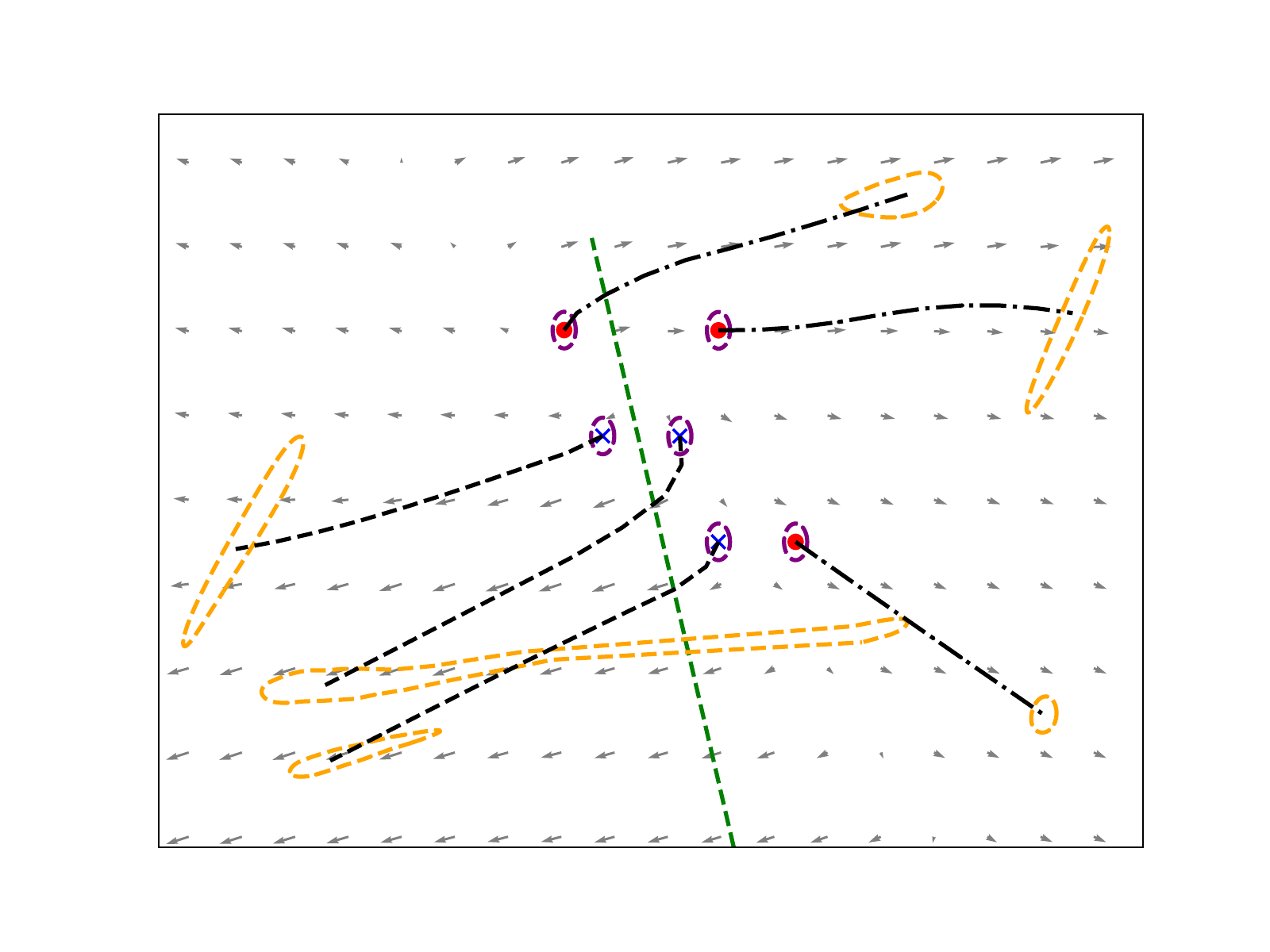} 
    \caption{Flows } 
    \label{fig7:b} 
    \vspace{0.5ex}
  \end{subfigure} 
  \begin{subfigure}[t]{0.5\linewidth}
    \centering
    \includegraphics[width=1.\linewidth]{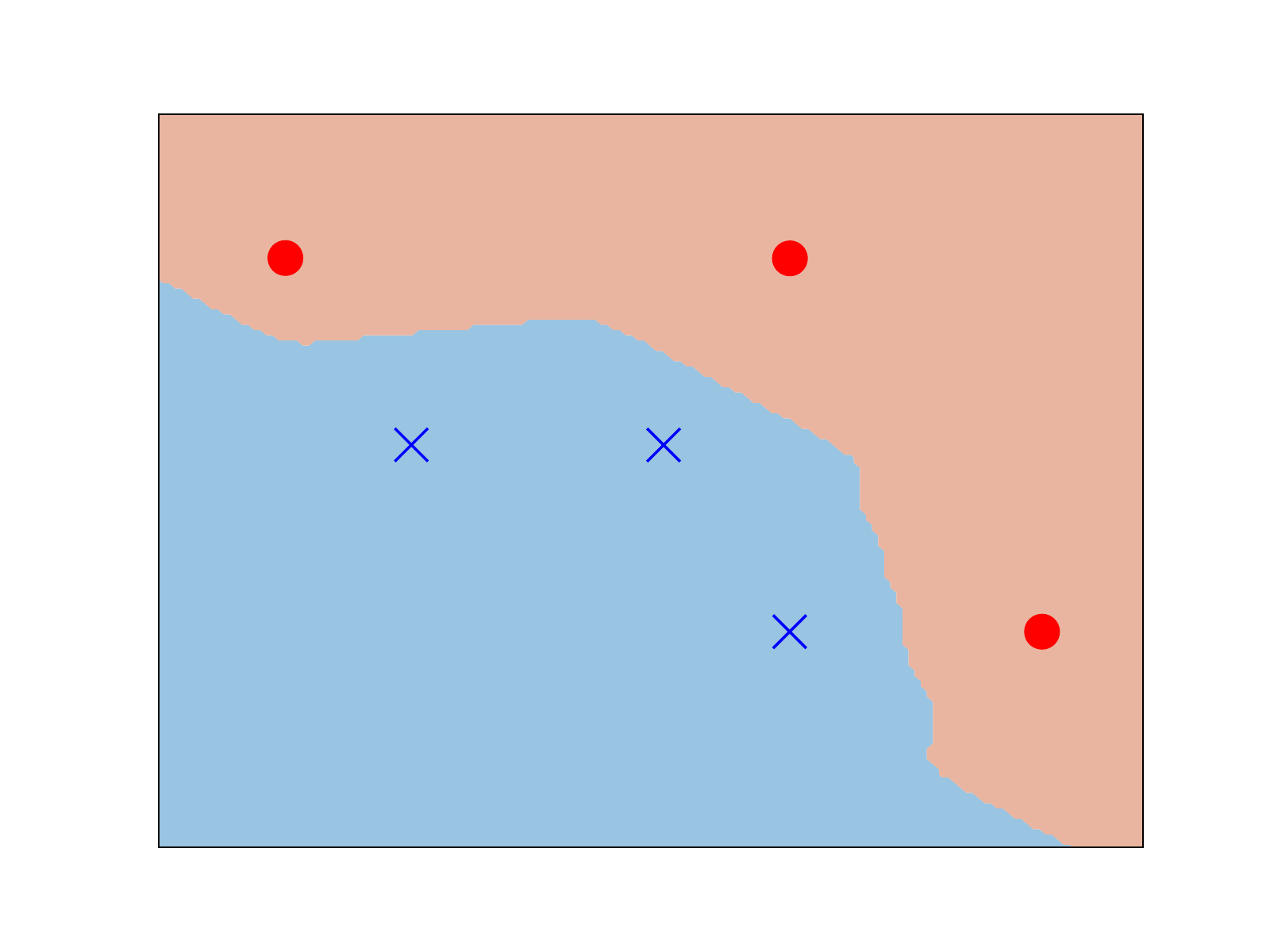} 
    \caption{Classification} 
    \label{fig7:c} 
  \end{subfigure}
  \begin{subfigure}[t]{0.5\linewidth}
    \centering
    \includegraphics[width=1.\linewidth]{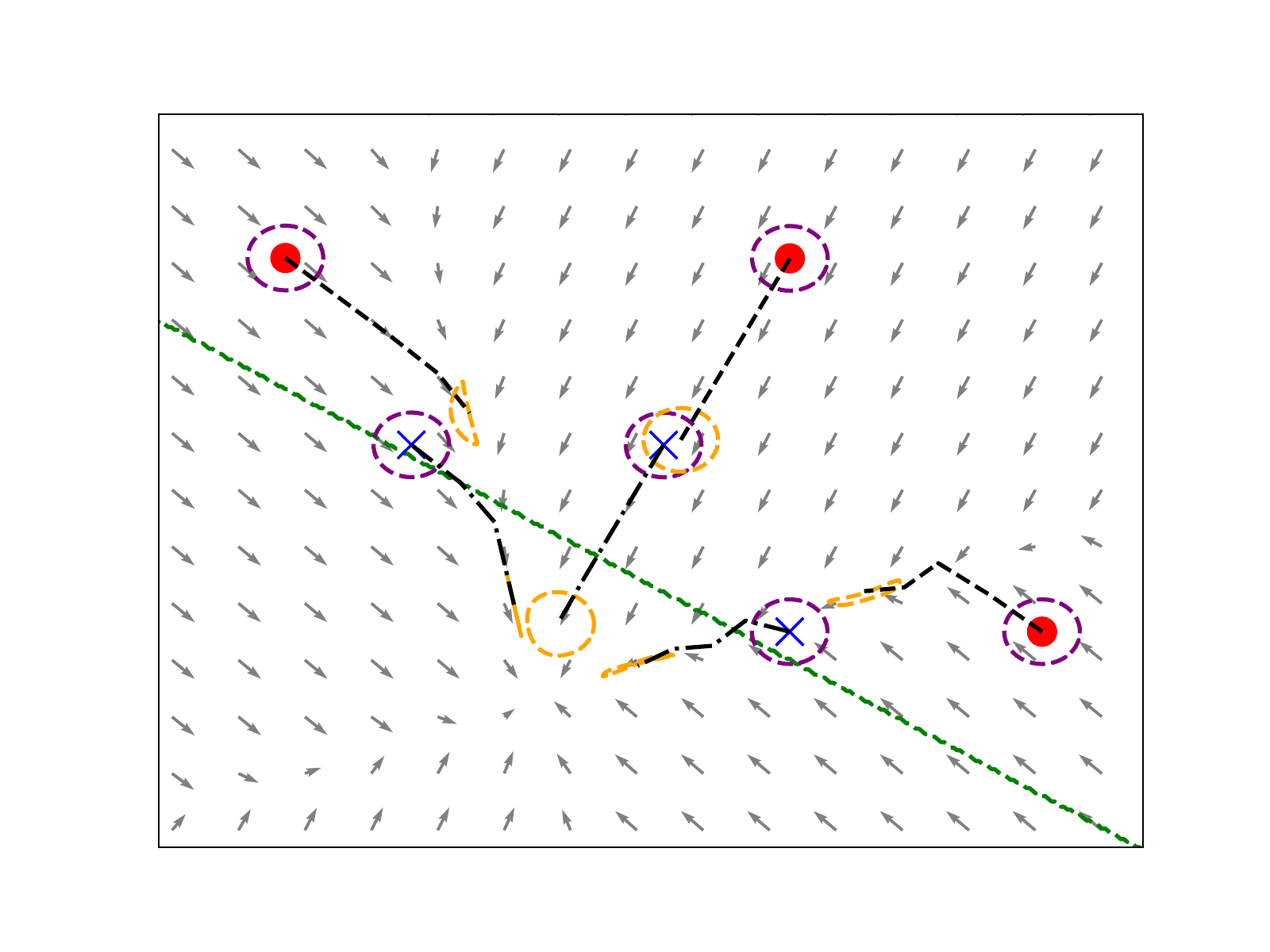} 
    \caption{Flows} 
    \label{fig7:d} 
  \end{subfigure} 
  \caption{Comparison between a vanilla NODE (panels (a)-(b)) and a CH-NODE (panels (c)-(d)) on a 2D binary classification task. The quiver plot shows the flows of the trained NODEs. Panels (b) and (d) show the propagation of the training data (black dashed lines) as well as how perturbed data (purple circles of radius $\epsilon_r$) are propagated in the output space (yellow sets). Dashed green lines: classification hyperplanes in the output space.}
  \label{fig:comp} 
\end{figure}

To show how contractivity promotes robustness, we compared a vanilla NODE and a time-invariant CH-NODE\footnote{Setting $\bm{K}(t) = \bm{K}$, $\bm{b}(t) = \bm{b}$, $\bm{L}(t) =0$ as constant matrices in \eqref{CH_NODE}} on a 2D binary classification task. The training dataset composed of 3 points for each class is shown in Fig. \ref{fig:comp}.
By using a linear \emph{softmax} classifier as the output layer, we trained both NODEs to correctly classify red and blue points.
The classification results for the vanilla NODE and the CH-NODE are illustrated in Fig. \ref{fig7:a}, and Fig. \ref{fig7:c}, respectively, where the colored regions represent the predictions of the trained NODEs.
It is clear from Fig. \ref{fig7:a} that the blue training points are closer to the classification boundary compared to Fig. \ref{fig7:d}.
Therefore, a small perturbation of the training set can result in wrong predictions for the vanilla NODE.

This is confirmed by Fig. \ref{fig7:b}, where small perturbations of the blue points in the middle give rise to an elongated output set intersecting the output classification hyperplane and, therefore, generating wrong predictions. However, as shown in Fig. \ref{fig7:d}, the problem disappears for the CH-NODE. In order to quantify the degree of robustness of both NODEs, we computed the maximal radius $\epsilon_r$ of the input perturbation balls (purple circles in Fig. \ref{fig7:b} and \ref{fig7:d}) still guaranteeing perfect classification. 
We obtained $\epsilon_r = 0.21$ for the vanilla NODE and for the CH-NODE $\epsilon_r = 0.7$. This confirms that the CH-NODE is more robust.


\subsection{MNIST Classification}

We consider the digit recognition problem using the MNIST \cite{deng2012mnist} dataset, and compare the robustness of the proposed NODEs with other standard neural networks.
Following \cite{RN10726}, and \cite{RN11564}, we employ a network architecture consisting of: i) a convolutional layer (input layer), followed by ii) NODE (continuum of hidden layer ), and iii) a \emph{softmax} output layer.

We compare the performance of the CH-NODE with the ResNet and the H-DNN proposed in \cite{RN11564}, which has a similar structure to the CH-NODE. 
For the implementation, we used Forward Euler discretization \eqref{eq:FE} with a step size of {$h=0.1$}. Moreover, to have comparable numbers of trainable parameters in all networks, we set $\bm{L} =0$ and $\kappa = 4 \times 10^{-4}$ in \eqref{energy_func}. We obtain the following NN architectures
\begin{equation*}
\fontsize{9}{9}
\begin{array}{lll}
    \text{ResNet:} \quad  \bm{\xi}_{i+1} = \bm{\xi}_{i} + \tanh{(\bm{K}_i \bm{\xi}_{i} + \bm{b}_i}) \\ 
        \text{H-DNN:} \quad  \bm{\xi}_{i+1} = \bm{\xi}_{i} + h \bm{J}\bm{K}^\top_i \tanh{(\bm{K}_i\bm{\xi}_{i} + \bm{b}_i}) \\ 
             {\text{CH-NODE:}} \hspace{-2mm} \quad  \bm{\xi}_{i+1} = (\bm{I}+ \kappa h\bm{F})\bm{\xi}_{i} + h \bm{F}\bm{K}^\top_i \tanh{(\bm{K}_i\bm{\xi}_{i} + \bm{b}_i})  
\end{array}
\end{equation*}
for $i = 0, 1, \cdots, N-1$, where $N$ is the number of layers, $\bm{F} = \bm{J} - \gamma \bm{I}$ and $\bm{J} = \begin{bmatrix}
\bm{0} & \bm{I} \\ -\bm{I} & \bm{0}
\end{bmatrix}$ 
has suitable dimensions.

\begin{figure}
    \centering
    \includegraphics[width = 0.65\linewidth]{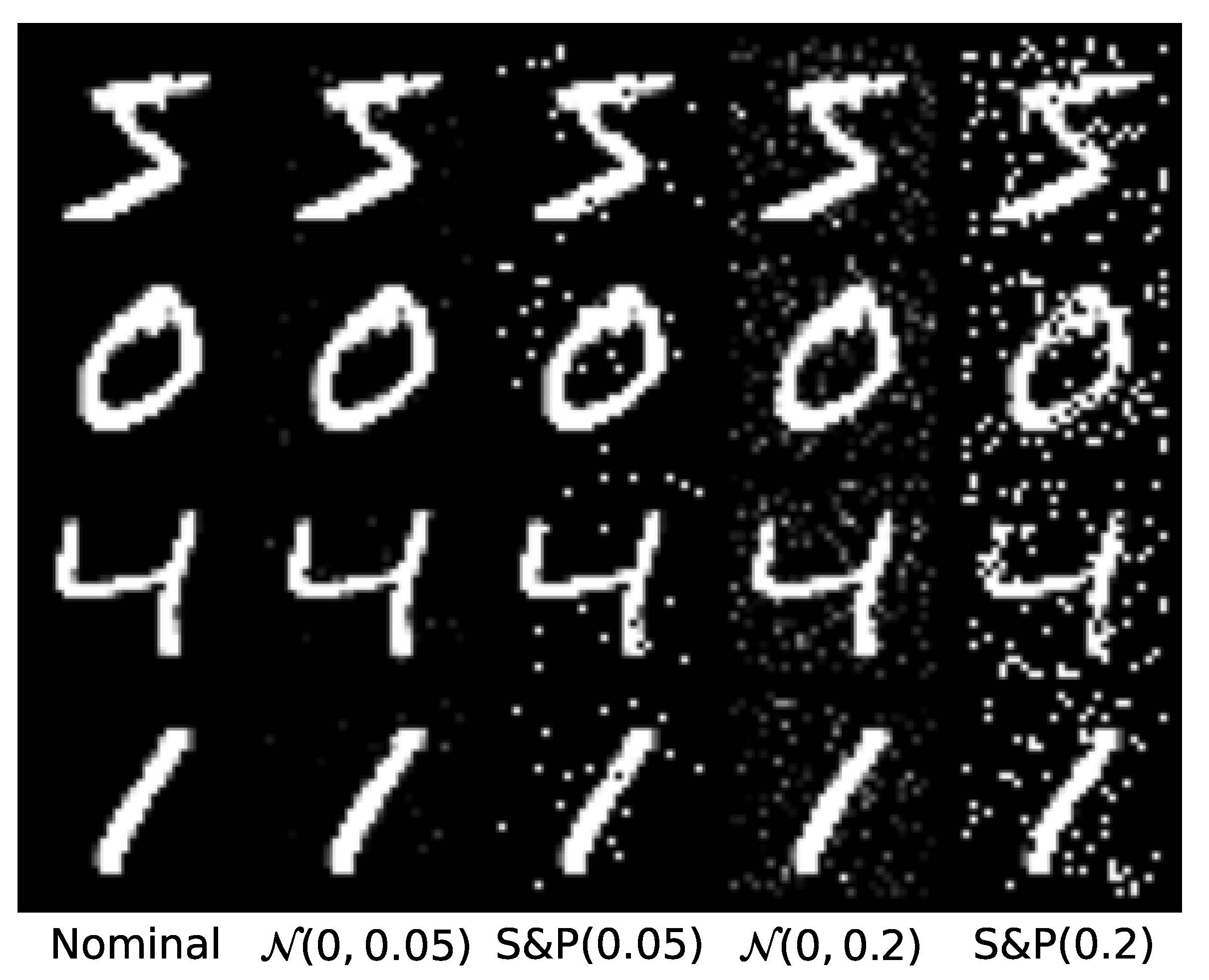}
    \caption{Corrupted MNIST test samples for the classification}
    \label{fig:MNIST}
\end{figure}

The robustness of NNs  can be evaluated through
the lens of their performance on perturbed images. Thus, we corrupted the test dataset with two types of noises of various magnitudes to compare the robustness. Specifically, we chose zero-mean Gaussian noise $\mathcal{N}(0, \sigma)$, and the salt and pepper noise $s \& p(\sigma)$ with $\sigma = [0.05, 0.2]$
\cite{schott2018towards}. 
For the Gaussian noise $\sigma$ denotes the variance, and for $s\&p$, $\sigma$ denotes the percentage  of corrupted pixels with the impulse noise (white or black pixel). Thus, we have four different datasets: a few samples from each one are shown in Fig. \ref{fig:MNIST}. 

{The trained networks were evaluated against noisy datasets in four different experiments.
In each experiment, ten noisy test datasets, each with $10000$ samples, were used. The corresponding average accuracy for each architecture is given in Table \ref{Table:1}. }
We can see that, in all cases, the test accuracy for ResNet and H-DNNs reduces significantly for noisy test points. On the other hand, {CH-NODEs} perform much better in {all} noisy scenarios. This supports our claim, that indeed, contractivity enhances robustness against small perturbations. 

\begin{table}
\centering
\resizebox{\columnwidth}{!}{%
\begin{tabular}{|c|c|c|c|c|c|c|c|}
    \hline
    & \multicolumn{1}{c|}{}
    &   \multicolumn{2}{c|}{\text{Nominal}}
    &   \multicolumn{2}{c|}{\textbf{$\mathcal{N}(0,\sigma)$}}
        &   \multicolumn{2}{c|}{\textbf{$s\&p(\sigma)$}}     \\  
     \hline
\text{N} &
 \text{NN}
        &    \text{Train}
            &   \text{Test}
        &    \textbf{{$\sigma = 0.05$}}
            &   \textbf{$\sigma = 0.2$}
        &    \textbf{$\sigma = 0.05$}
            &   \textbf{$\sigma = 0.2$}\\  
 \hline
\multirow{3}{0.75em}{\textbf{4}} & \text{ResNet} & \textbf{98.91} & \textbf{97.01} &63.00	&52.56 	&59.8	&42.02             \\ 
& \text{H-DNN}  & 94.68 & 94.60 & 31.12 &26.65& 30.52 &	23.83              \\
& \text{CH-NODE} & 94.03  & 92.38 & \textbf{81.30}	& \textbf{77.69} &	\textbf{79.86}&	\textbf{63.84}            \\
\hline
\multirow{3}{0.75em}{\textbf{8}}& \text{ResNet}  & \textbf{99.12} & \textbf{97.28} &32.99	&30.56	&30.27	&28.11         \\  
& \text{H-DNN}  & 95.30 & 95.17 & 60.8 &	49.88 &	61.15 &	45.62              \\
& \text{CH-NODE}  & 89.55 &  89.01 & \textbf{86.33}	&\textbf{81.85}&	\textbf{84.22} &	\textbf{72.18}             \\
\hline
\multirow{3}{0.75em}{\textbf{12}}& \text{ResNet}  & \textbf{99.11} & \textbf{96.86} & 39.13 & 34.04	& 41.04	& 29.80          \\  
& \text{H-DNN}  & 95.36 & 95.23 &  26.79	& 23.53 &	27.48 &	22.75            \\
& \text{CH-NODE} & 90.01 &  89.76 & \textbf{85.68}	& \textbf{80.97} &	\textbf{84.88}&	\textbf{72.82}              \\
\hline
\end{tabular}%
}
 \caption{Robustness comparison between ResNets, H-DNNs, and {CH-NODEs} under the zero-mean Gaussian and the salt and pepper noise. For each value of $N$, the best performance in each column appears in \textbf{bold}.}
  \label{Table:1}
 \end{table}

\subsection{Numerical validation of non-exploding gradients}
\begin{figure}
    \centering
    \includegraphics[width = \linewidth]{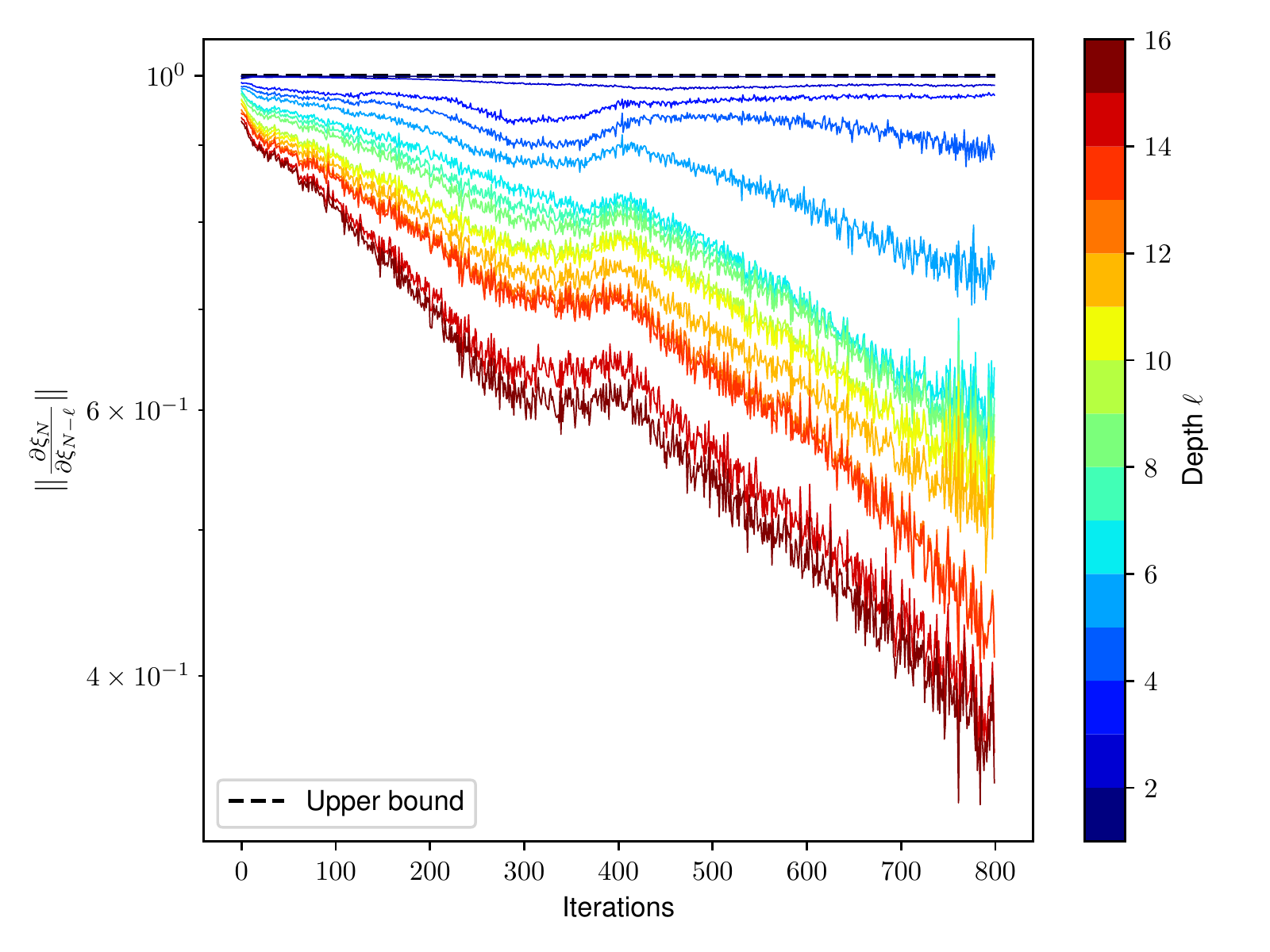}
    \caption{Evolution of the 2-norm of the BSM during the training of a 16-layer {CH-NODE} exhibiting non-exploding gradients.}
    \label{fig:norms_grad}
\end{figure}

We demonstrate the non-exploding gradients property of CH-NODEs on the “Double circles” dataset used in \cite{RN11564}. We discretize the ODE \eqref{CH_NODE} with $\tanh{(\cdot)}$ activation function using FE method with a step-size of $h = 6.25 \times 10^{-4}$ to obtain a {CH-NODE} with 16 layers, each with inputs and outputs in $\mathbb{R}^2$. We set $\bm{L} = 0$ and $\kappa = 0.04$ in \eqref{CH_NODE}, and choose $\epsilon = 1.0 \times 10^{-9}$. The norms of the BSMs in each training iterations are displayed in Fig. \ref{fig:norms_grad}. It is easy to see that the norm of BSMs remains bounded between $0.3214 \leq ||\frac{\partial \bm{\xi}(T)}{\partial \bm{\xi}(T - t)}|| \leq 1$.  Moreover,  the NODE achieves a train accuracy of $99.20 \%$ and a test accuracy of $99.12 \%$. 

\section{CONCLUSIONS}
We proposed NODEs based on Hamiltonian dynamics that are contractive by design.
Furthermore, we have shown that, CH-NODEs cannot suffer from exploding gradients. The robustness of these networks against input noise has been validated using benchmark classification examples. 
Further research will be devoted to analyze the robustness of CH-NODEs against different forms of adversarial attacks \cite{RN11056}, and explore discretization schemes that preserve contractivity.

\section{ACKNOWLEDGEMENTS}
The authors would like to acknowledge the useful discussions with Corentin Briat.

\section{APPENDIX}
\subsection{Proof of Theorem \ref{prop-1}}
\label{appen1}
The Hessian of the Hamiltonian function~\eqref{energy_func} is
\begin{align*}
  \frac{\partial^2 H(\bm{\xi},t) }{\partial \bm{\xi}^2} 
    &= \bm{K}^\top(t) \bm{D}(\bm{\xi},t) \bm{K}(t) + \bm{L}^\top(t) \bm{L}(t) +\kappa \bm{I} \;,
\end{align*}
where {$\bm{D}(\bm{\xi},t) = \text{diag}(\sigma' (\bm{K}(t)\bm{\xi} + \bm{b}(t)))$} is a diagonal matrix.
By exploiting the properties of {$\sigma'(\cdot)$}, we can bound this diagonal matrix as {$\bm{0} \preceq \bm{D}(\bm{\xi},t) \preceq S\bm{I}$}.
Therefore, we have
\begin{align}
\label{Hess_cond}
  \bm{0} \prec c_1 \bm{I} \preceq \frac{\partial^2 H(\bm{\xi},t) }{\partial \bm{\xi}^2} \preceq c_2 \bm{I} \; ,
\end{align}
where the constants $c_1$ and $c_2$ are defined in~\eqref{eq.c1Definition} and~\eqref{eq.c2Definition}, respectively.
Denote $\bar{\lambda}_J = \bar{\lambda}(\bm{J} \bm{J}^\top)$, if $\gamma$ is selected to satisfy~\eqref{eq:gammaCondition}, after some manipulations, we have
\begin{align}
\label{first_step}
-\frac{\gamma^2}{(\gamma^2+ \bar{\lambda}_J)^2}+\frac{\alpha^2+\epsilon}{\gamma^2+ \bar{\lambda}_J} \leq 0 \;.
\end{align}
Let $\nu=\frac{\gamma}{\gamma^2+ \bar{\lambda}_J}>0$, and add $(\nu - \frac{\gamma}{\gamma^2+\bar{\lambda}_J})^2$ to the LHS of \eqref{first_step} to obtain
\begin{align*}
  \left(\nu-\frac{\gamma}{\gamma^2+\bar{\lambda}_J} \right)^2-\frac{\gamma^2}{(\gamma^2+\bar{\lambda}_J)^2}+\frac{\alpha^2+\epsilon}{\gamma^2+\bar{\lambda}_J} \le 0 
\end{align*}
which, after some manipulations, gives
$\nu^2 (\gamma^2+\bar{\lambda}_J)-2\nu\gamma +\alpha^2+\epsilon \le 0.
$
Since $\bm{J} \bm{J}^\top \preceq \bar{\lambda}_{J} \bm{I}$, we further have 
\begin{align}
\label{step_int}
    \begin{aligned}
         -2 \nu \gamma \bm{I}  + \alpha^2 \bm{I} + \epsilon \bm{I} +  \nu^2 \bm{J} \bm{J}^\top + \gamma^2  \nu^2 \bm{I}  \preceq 0 \; . 
\end{aligned}
\end{align}
By exploiting that $\bm{J} + \bm{J}^\top = 0$, it is straightforward to show that \eqref{step_int} is equivalent to 
$\nu( \bm{J} - \gamma \bm{I} ) + \nu ( \bm{J} - \gamma \bm{I} )^\top + \alpha^2 \bm{I} + \epsilon \bm{I}  
          + \nu^2 ( \bm{J} - \gamma \bm{I} ) ( \bm{J} - \gamma \bm{I} )^\top \preceq 0.$
By defining $\bm{F} = \bm{J} - \gamma \bm{I}$ and $\bm{P} = \nu \bm{I}$, we can write the above inequality as 
\begin{equation}
\label{lmi_schur}
    \bm{P} \bm{F} + \bm{F}^T \bm{P} + \alpha^2 \bm{I} + \epsilon \bm{I} + \bm{P} \bm{F} \bm{F}^\top \bm{P} \preceq 0 \;,
\end{equation}
which is equivalent to the following inequality by using the Schur complement lemma
\begin{align}
\label{LMI}
    \begin{bmatrix}
    \bm{P} \bm{F} + \bm{F}^T \bm{P} + \alpha^2 \bm{I} + \epsilon \bm{I} & \bm{P} \bm{F} \\ 
    \star & - \bm{I} 
    \end{bmatrix} \preceq 0 \;. 
\end{align}
Then in view of Proposition 3 in~\cite{barabanov2019contraction} and~\eqref{Hess_cond}, we conclude that the Hamiltonian system~\eqref{CH_NODE} is contractive {with a contraction rate $\epsilon$}. 

\subsection{Proof of Theorem \ref{Thm:2}}
\label{Appen2}
Define $\bm{\Phi}(T,T-t) = \frac{\partial \bm{\xi}(T)}{\partial \bm{\xi}(T - t)}$, following the derivations in~\cite[Appendix II]{RN11564}, we have
\begin{align}
\label{dyn_back}
 \dot{\bm{\Phi}}(T,T-t) = \bm{A}(T -t) \bm{\Phi}(T,T-t) \;,
\end{align}
where $t \in [0, \ T]$ and 
\begin{align}
\label{eq-proof1}
\begin{aligned}
     \bm{A}(T-t) =  \frac{\partial^2 H(\bm{\xi}(T-t), T- t) }{\partial \bm{\xi}^2(T-t)}  \left( \bm{J} - \gamma \bm{I} \right)^\top \;.  
\end{aligned}
\end{align}
For the sake of brevity, we denote $\bm{\Phi} := \bm{\Phi}(T,T-t)$, $$\displaystyle{\nabla^2 H := \frac{\partial^2 H(\bm{\xi}(T-t), T- t) }{\partial \bm{\xi}^2(T-t)}}$$ and further define 
\begin{align*}
    \bm{X}(T-t) := \nabla^2 H  - \beta \bm{I}, \ 
    \bm{\zeta}(T-t) = \bm{X}(T-t)\bm{\Phi} \;, 
\end{align*}
where $\beta = \frac{1}{2}(c_1 + c_2) $. 
Let $\bm{V}(T,T-t) = \nu \beta \bm{\Phi}^\top \bm{\Phi}$, where $\nu = \frac{\gamma}{\gamma^2+\bar{\lambda}_J}$ and denote $\bm{V} :=\bm{V}(T,T-t)$ as a shorthand notation. Then, we have the following differential equation along the trajectories of \eqref{dyn_back}
\begin{align}
    \dot{\bm{V}} &= \nu \beta( \bm{\Phi}^\top \dot{\bm{\Phi}} + \dot{\bm{\Phi}}^\top  \bm{\Phi} ) \label{lyap_diff} \\
    &= \nu \beta ( \bm{\Phi}^\top \nabla^2 H \bm{F}^\top \bm{\Phi} + \bm{\Phi}^\top \bm{F} \nabla^2 H \bm{\Phi}) \nonumber \\
     &=  \nu \beta ( \bm{\Phi}^\top (\bm{X}^\top + \beta \bm{I}) \bm{F}^\top  \bm{\Phi} +   \bm{\Phi}^\top \bm{F}(\bm{X}+ \beta \bm{I}) \bm{\Phi} ) \nonumber \\ 
     &= \nu \beta(\bm{\zeta}^\top  \bm{F}^\top  \bm{\Phi} + \beta  \bm{\Phi}^\top  \bm{F}^\top \bm{\Phi}  +    \bm{\Phi}^\top  \bm{F}  \bm{\zeta} + \beta  \bm{\Phi}^\top  \bm{F} \bm{\Phi} \nonumber
\end{align}
where we used the short hand notations $\bm{\zeta} := \bm{\zeta}(T-t)$, and $\bm{X} := \bm{X}(T-t)$. For bounding the norm of the BSMs, we have to show that $\dot{\bm{V}} \preceq - \delta \bm{\Phi}^\top \bm{\Phi}$, where $\delta > 0$. 
Note, from \eqref{Hess_cond}, one has 
$-\mu \bm{I} \preceq \bm{X}(T-t) \preceq \mu \bm{I},  \ \mu= \frac{1}{2}(c_2 - c_1)\;,$
which implies 
\begin{align}
\label{final_eq}
     \mu^2 \bm{\Phi}^\top \bm{\Phi} - \bm{\zeta}^\top \bm{\zeta} \succeq 0 \; .
\end{align}
Adding \eqref{final_eq} to RHS of $\dot{\bm{V}}$ yields the following inequality
\begin{align}
\dot{\bm{V}}  &\preceq   \mu^2 \bm{\Phi}^\top \bm{\Phi} - \bm{\zeta}^\top \bm{\zeta} \label{intermeditae} \\ 
& \hspace{-4mm}+ \nu \beta( \bm{\zeta}^\top  \bm{F}^\top  \bm{\Phi} + \beta  \bm{\Phi}^\top  \bm{F}^\top \bm{\Phi}  +    \bm{\Phi}^\top  \bm{F}  \bm{\zeta} + \beta  \bm{\Phi}^\top  \bm{F} \bm{\Phi} ) \; , \nonumber
\end{align}
and then adding a small positive term $\delta \bm{\Phi}^\top \bm{\Phi}$ to both sides \eqref{intermeditae}, we have 
\begin{align*} 
\begin{aligned}
\dot{\bm{V}}  + \delta \bm{\Phi}^\top \bm{\Phi}  &\preceq   \mu^2 \bm{\Phi}^\top \bm{\Phi} - \bm{\zeta}^\top \bm{\zeta}  + \delta \bm{\Phi}^\top \bm{\Phi} \\ 
& \hspace{-9mm}+ \nu \beta( \bm{\zeta}^\top  \bm{F}^\top  \bm{\Phi} + \beta  \bm{\Phi}^\top  \bm{F}^\top \bm{\Phi}  +    \bm{\Phi}^\top  \bm{F}  \bm{\zeta} + \beta  \bm{\Phi}^\top  \bm{F} \bm{\Phi} ) \; .
\end{aligned}
\end{align*}
After some calculations, we obtain 
\begin{align}
\begin{aligned}
\label{int_lmi}
\dot{\bm{V}} + \delta \bm{\Phi}^\top \bm{\Phi} &\preceq  \\
&\hspace{-15mm}\begin{bmatrix}
\bm{\Phi} \\ \bm{\zeta}
\end{bmatrix}^\top \hspace{-1mm}
      \begin{bmatrix}
    \nu  \beta^2 \bm{F} + \nu \beta^2 \bm{F}^T  + \mu^2 \bm{I} + \delta \bm{I} & \nu \beta \bm{F} \\ 
    \star & - \bm{I} 
    \end{bmatrix}
    \begin{bmatrix}
\bm{\Phi} \\ \bm{\zeta}
\end{bmatrix}.
\end{aligned}
\end{align}
By multiplying \eqref{int_lmi} from the left and from the right by $\text{diag}(\bm{I}/\beta, \bm{I})$, we obtain the left hand side of \eqref{LMI} with $\alpha^2 = \mu^2 / \beta^2$, $\epsilon = \delta / \beta^2$, and $\bm{P} = \nu \bm{I}$. 
Since according to Theorem \ref{prop-1}, the LMI \eqref{LMI} holds by design for CH-NODE, $\dot{\bm{V}} + \delta \bm{\Phi}^\top \bm{\Phi} \preceq 0$ also holds and implies $\dot{\bm{V}} \preceq  -\delta \bm{\Phi}^\top \bm{\Phi}$. Moreover, one can easily show that 
\begin{align*}
    \dot{\bm{V}} \preceq - \delta \bm{\Phi}^\top \bm{\Phi} =  -(\frac{\delta}{\nu \beta }) \nu \beta \bm{\Phi}^\top \bm{\Phi} = -(\frac{\epsilon \beta }{ \nu }) \bm{V}  \preceq -  \rho \bm{V},
\end{align*}
where $\rho = \frac{\epsilon \beta (\gamma^2 + \bar{\lambda}_J)}{ \gamma }$. {It is straightforward to verify that $\bm{V} \preceq e^{-\rho t}\bm{V}(T,T)$, and since  $\bm{\Phi}(T,T) = \bm{I}$, $\bm{\Phi}^\top \bm{\Phi} - e^{-\rho t} \bm{I} \preceq 0$ also holds. 
Finally, the maximum singular value of $\bm{\Phi}$ can be bounded as $\sigma_{\max}(\bm{\Phi}) \leq e^{-\frac{\rho}{2} t}$, and hence, $||\bm{\Phi}(T,T-t)|| \leq e^{- \frac{\rho}{2} t} \leq 1$ holds. }

\bibliography{bibliography}
\bibliographystyle{IEEEtran}

\end{document}